\documentclass{article}
\usepackage{ijcai21}
\usepackage{times} 
\usepackage[hidelinks]{hyperref}
\usepackage[utf8]{inputenc}
\usepackage[small]{caption}
\usepackage{graphicx}
\usepackage{amsmath,amssymb}
\usepackage{amsthm}
\usepackage{booktabs,multirow}
\usepackage{siunitx}
\usepackage{color,hyphenat,balance,booktabs,microtype,anyfontsize}
\usepackage[fleqn,tbtags]{mathtools}
\usepackage{overpic}
\usepackage[switch]{lineno}
\usepackage[capitalise]{cleveref}
\graphicspath{{./images/}}

\def \etal {{\emph{et al}.\thinspace}}

\def \ie {{\emph{i.e}.\thinspace}}
\def \mx {\mathbf{x}}
\def \mn {\mathbf{n}}

\newcommand{\tablestyle}[2]{\setlength{\tabcolsep}{#1}
                            \renewcommand{\arraystretch}{#2}
                            \centering\footnotesize}

\title{Spline Positional Encoding for Learning 3D Implicit Signed Distance Fields}

\author{
  Peng-Shuai Wang$^1$ \and
  Yang Liu$^1$        \and 
  Yu-Qi Yang$^{2,1}$  \and
  Xin Tong$^1$
\affiliations
  $^1$Microsoft Research Asia \\
  $^2$Tsinghua University
\emails
  \{penwan, t-yuqyan, yangliu, xtong\}@microsoft.com
}

\begin{document}

\maketitle


\begin{abstract}
Multilayer perceptrons (MLPs) have been successfully used to represent 3D shapes implicitly and compactly, by mapping 3D coordinates to the corresponding signed distance values or occupancy values.
In this paper, we propose a novel positional encoding scheme, called \emph{Spline Positional Encoding}, to map the input coordinates to a high dimensional space before passing them to MLPs, for helping to recover 3D signed distance fields with fine-scale geometric details from unorganized 3D point clouds.
We verified the superiority of our approach over other positional encoding schemes on tasks of 3D shape reconstruction from input point clouds and shape space learning.
The efficacy of our approach extended to image reconstruction is also demonstrated and evaluated.
\end{abstract}

\section{Introduction}
\label{sec:intro}

Implicit neural representations learned via multilayer perceptrons (MLPs) have been proved to be effective and compact 3D representations~\cite{Park2019,Mescheder2019,Chen2019} in computer vision and graphics fields.
The MLPs take 3D coordinates as input directly, denoted by \emph{coordinate-based MLPs}, and output the corresponding signed distance values or the occupancy values.
They essentially define continuous implicit functions in 3D space whose zero level set depicts shape surfaces.
Compared with conventional 3D discrete representations like point clouds or voxels, the MLP-based implicit representation has infinite resolutions due to its continuous nature, while being extremely compact.
Apart from representing 3D shapes, coordinate-based MLPs are also capable of representing images, 3D textures, and 5D radiance fields, serving as general-purpose mapping functions.

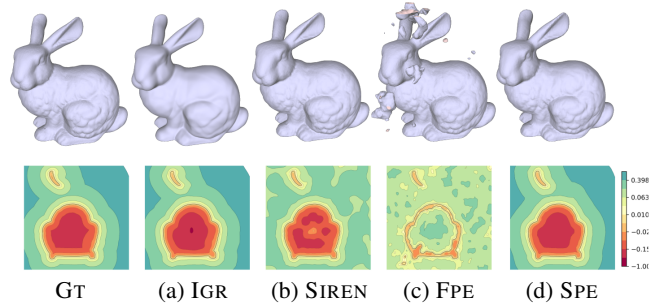
\begin{figure}
  \centering
  \begin{overpic}[width=1\linewidth]{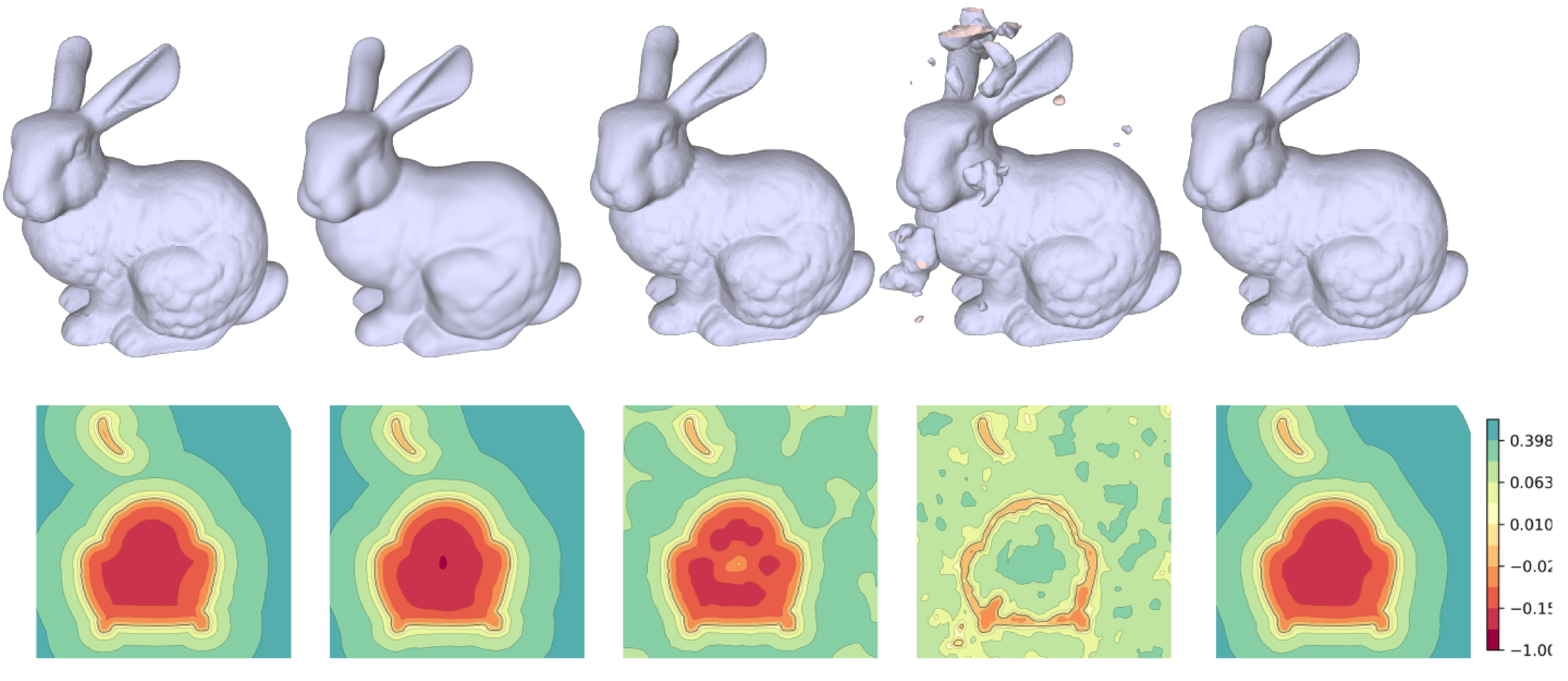}
    \put(7, -3){\small  \textsc{Gt}}
    \put(23, -3){\small (a) \textsc{Igr}}
    \put(40.5, -3){\small (b) \textsc{Siren}}
    \put(61, -3){\small (c) \textsc{Fpe}}
    \put(80, -3){\small (d) \textsc{Spe}}
  \end{overpic}
  \vspace{0.1mm}
  \caption{SDF learning via MLP-based methods. Upper: the extracted zero level
    set via marching cubes. Lower: a slice view of SDFs. \textsc{Siren}~\protect\cite{Sitzmann2020},
    \textsc{Fpe}~\protect\cite{Tancik2020} and our \textsc{Spe} fit the input point coordinates and
    normals well,  but \textsc{Fpe} contains many unwanted small branches. The
    SDFs from \textsc{Igr}~\protect\cite{Gropp2020} and \textsc{Spe} are more faithful to the
    ground-truth, while \textsc{Spe} recovers more details. }
  \label{fig:teaser}
\end{figure}

In this paper, we are interested in learning the signed distance field (SDF) effectively from an unorganized input point cloud sampled from a 3D shape, by MLPs.
SDF defines the distance of a given point $\mx$ from the shape surface, with the sign determined by whether $\mx$ is inside the shape volume or not.
SDFs are needed by a broad range of applications~\cite{Jones2006}, including, but not limited to, 3D reconstruction~\cite{Curless1996}, constructive solid geometry (CSG), collision detection~\cite{Bridson2015} and volume rendering~\cite{Hart1996}.
The recent MLP-based approaches~\cite{Gropp2020,Atzmon2020} introduce the Eikonal equation constraint $|\nabla F(\mx)|\equiv 1$ to the mapping function $F: \mx \in \mathbb{R}^3$ to enforce $F$ to be an SDF, while its zero level set passes through the point cloud.
However, due to the ``spectral bias'' of neural networks~\cite{Rahaman2019,xu2019training,Mildenhall2020,Tancik2020}, coordinate-based MLPs with ReLU activation are incapable of reconstructing high-frequency details of surfaces.
An example produced by a coordinate-based MLP --- \textsc{Igr}~\cite{Gropp2020} is shown in \cref{fig:teaser}(a), where the output shape is over-smoothed compared to the ground-truth.

To circumvent this problem, \textsc{Siren}~\cite{Sitzmann2020} uses Sine as the activation function in place of ReLU to improve the expressiveness of MLPs, and the Fourier Positional Encoding (abbreviated as \textsc{Fpe})~\cite{Mildenhall2020,Tancik2020,Zhong2020} is proposed to improve network capability by lifting input coordinates to a high-dimensional Fourier space via a set of sinusoidal functions before feeding the coordinates as the input of MLPs.
However, both approaches fail to recover SDFs in good quality and are even worse than \textsc{Igr} (see \cref{fig:teaser}(b)\&(c)), although their zero level sets may fit the point cloud well.


In this paper, we propose a novel \emph{Spline Positional Encoding} (abbreviated as \textsc{Spe}), with which the MLP can not only recover the high-frequency details of the surface but also recover the SDF well, as shown in \cref{fig:teaser}(d).
Our \textsc{Spe} maps the input coordinates into a high-dimensional space via projecting them onto multiple trainable Spline functions, instead of hard-coded sinusoidal functions as \textsc{Fpe}.
The Spline functions are defined as the weighted sum of a series of uniformly spaced local-support B-Spline bases, and the weights are trainable.
As the Spline function can be used to approximate other continuous functions, our \textsc{Spe} can be regarded as a generalization of \textsc{Fpe}.
\textsc{Spe} greatly increases the fitting ability of MLPs to reproduce high-frequency details.
By subdividing the B-Spline bases, \textsc{Spe} can also be progressively refined.
Based on this property, we also design a multi-scale training scheme to help
MLPs converge to better local minima, which enables our network to recover SDFs and geometric details progressively and robustly.

Through experiments and ablation studies, we demonstrate the efficacy and the superiority over other state-of-the-art encoding schemes of our \textsc{Spe} on the tasks of learning SDFs from a point cloud or a set of point clouds.
Additionally, to test the generalizability of \textsc{Spe}, we also apply it to image reconstruction and achieve good performance.

\section{Related Work}
\label{sec:related}

\paragraph{Coordinate-based MLPs.}
The coordinate-based MLPs have caught great research interest as a continuous representation of shapes~\cite{Park2019,Mescheder2019,Chen2019}, scenes~\cite{Sitzmann2019}, images~\cite{Tancik2020,Sitzmann2020}, textures~\cite{Oechsle2019} and 5D radiance fields~\cite{Mildenhall2020}.
These methods train MLPs by regressing the ground truth SDFs, point/pixel colors, or volume radiance values.
Our work is motivated by the works~\cite{Atzmon2020a,Atzmon2020,Gropp2020} that use MLPs to reconstruct SDFs from raw point clouds, without knowing the ground truth SDFs.

The limitation of coordinate-based MLPs with ReLU activation has been revealed by~\cite{Rahaman2019,Mildenhall2020}: the high-frequency fitting error decreases exponentially slower than the low-frequency error.
To overcome this issue, there are multiple attempts to improve the representation power of MLPs as follows.

\paragraph{Activation function.}
\textsc{Siren}~\cite{Sitzmann2020} use Sine as the activation function and proposes a proper initialization method for training.
It greatly improves the expressiveness of MLPs, and it is capable of recovering fine geometry details in the 3D reconstruction task.
However, ReLU can provide strong implicit regularization when being under-constrained~\cite{Gropp2020} and offer a good approximation to SDF in the whole space.
In our work, we choose Softplus as our activation function, which can be regarded as a differentiable ReLU.

\paragraph{Positional encoding.}
Sinusoidal encoding is a kind of positional encoding that is first used for representing 1D positions in natural language processing~\cite{Vaswani2017}.
This type of positional encoding has proved to be able to improve the performance of MLPs in radiance fields fitting~\cite{Mildenhall2020} and 3D protein structure reconstruction~\cite{Zhong2020}.
Tancik \etal~\shortcite{Tancik2020} build a theoretical connection between Sinusoidal mapping and Neural Tangent Kernels~\cite{Jacot2018} for proving the efficacy of sinusoidal mapping and further improve its performance by using random Fourier features~\cite{Rahimi2018}.
Their Fourier Positional Encoding (\textsc{Fpe}) maps input points to a higher dimensional space with a set of sinusoids.
However, \textsc{Fpe} is not suitable to minimize the loss function containing function gradient constraints as we reveal in \cref{sec:result}.

\paragraph{Local MLPs.}
Local MLPs improve the performance of a global MLP by dividing complex shapes or large-scale scenes into regular subregions~\cite{Peng2020,Chabra2020,Jiang2020,Genova2020} and fitting each subregion individually with the consideration of fusing the local output features or local output results.
Our Spline Positional Encoding is composed of uniformly spaced locally supported basis functions along with different project directions.
It shares the same sprite to local MLPs, but executes the local representation at the beginning of MLP.

\section{Spline Positional Encoding}
\label{sec:method}

In this section, we first briefly review the loss functions for learning SDFs from a point cloud in \cref{subsec:learnsdf}, then introduce our spline positional encoding and its relations to prior arts in \cref{subsec:spe}, and our training scheme in \cref{subsec:training}.

\subsection{SDF Learning} 
\label{subsec:learnsdf}

Given a set of points with oriented normals sampled from the unknown surface $\mathcal{S}$ of a 3D shape, denoted by $\mathcal{X}=\{(\mx_i, \mn_i)\}_{i \in \mathcal{I}}$, the goal is to train an MLP $F(x)$ which represents the SDF of $\mathcal{S}$ and keeps $F(\mx_i) = 0,  \nabla F(\mx_i) = \mn_i$, $\forall i \in \mathcal{I}$. 
To ensure $F(x)$ is an SDF, an additional constraint from the Eikonal equation $\|\nabla F(x) \|=1$ is added as recommended by \cite{Gropp2020}. 
The final loss function is in the following form.
\begin{equation}
  \begin{aligned}
    L_{sdf} = & \sum_{i \in \mathcal{I}} (F(x_i) ^2 + \tau \, \| \nabla F(\mx_i) - \mn_i \|^2) +  \\
              & \lambda \, \mathbb{E}_x (\| \nabla F(\mx) \| - 1)^2.
  \end{aligned}
  \label{equ:loss}
\end{equation}
After training, $F(\mx)$ approximates the underlying SDF induced by the input point clouds, and the zero level set $F(\mx)=0$ approximates $\mathcal{S}$, which can be extracted as a polygonal mesh via Marching Cubes~\cite{Lorensen1987}.

\subsection{Spline Positional Encoding}
\label{subsec:spe}

\begin{figure}[t]
  \centering
  \begin{overpic}[width=0.7\linewidth]{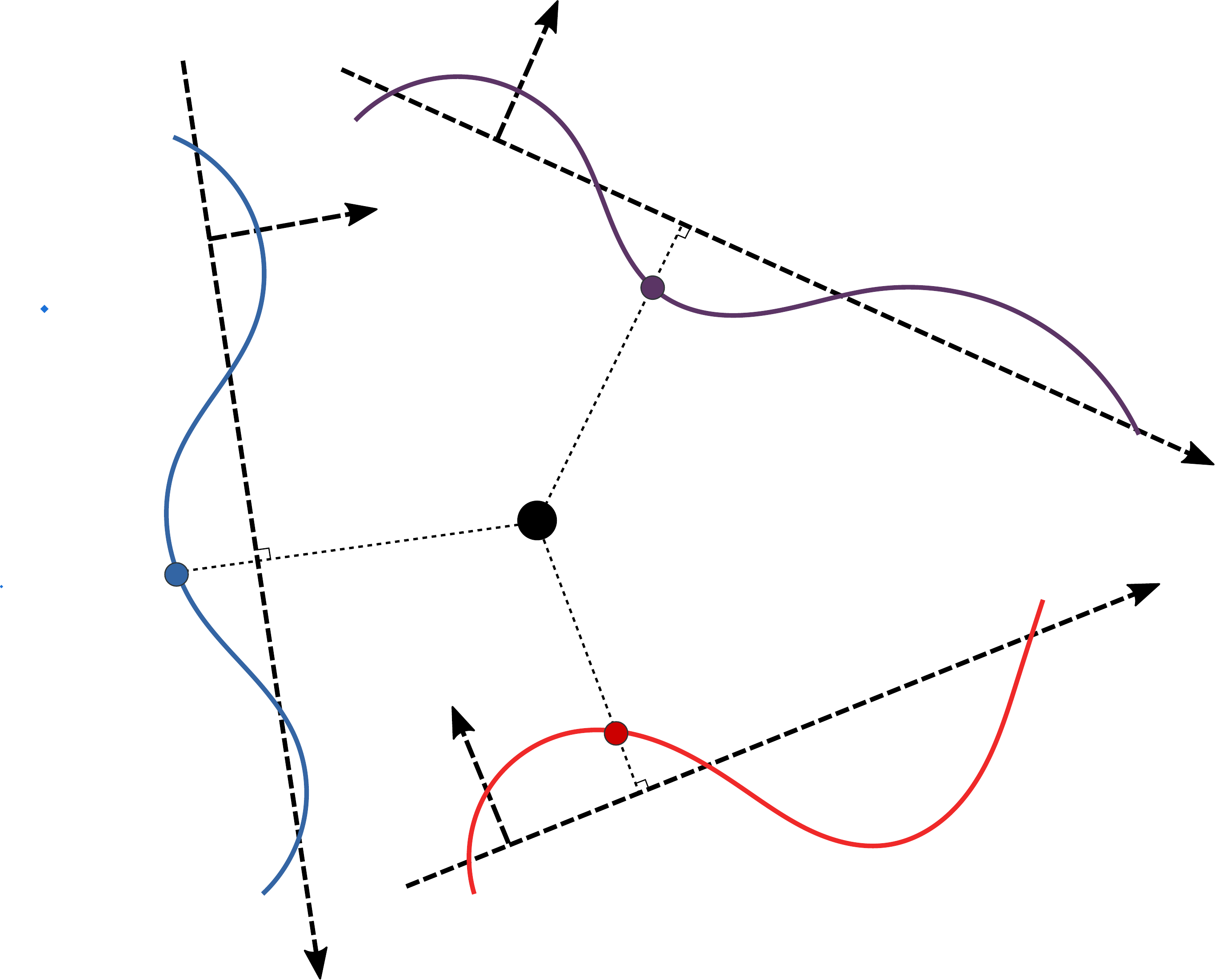}
    \put(47,36){\small $\mathbf{x}$} 
    \put(101,43){\small $x^\prime$}
    \put(48,76){\small $y^\prime$} 
    \put(94,34){\small $x^{\prime\prime}$}
    \put(39,21){\small $y^{\prime\prime}$} 
    \put(28,1){\small $x^{\prime\prime\prime}$} 
    \put(29,59){\small $y^{\prime\prime\prime}$}
    \put(53,51){\small $\mathbf{p}^\prime$} 
    \put(52,22){\small $\mathbf{p}^{\prime\prime}$} 
    \put(9,28){\small $\mathbf{p}^{\prime\prime\prime}$}
  \end{overpic}
  \caption{Illustration of Spline positional encoding on 2D. Point $\mx$ is
  projected onto three Splines along three directions. The local heights of
  $\mathbf{p}^\prime$,  $\mathbf{p}{^\prime\prime}$,
  $\mathbf{p}^{\prime\prime\prime}$ with respect to their own y-axis:
  $y^\prime$, $y^{\prime\prime}$, $y^{\prime\prime\prime}$ form the Spline
  positional encoding of $\mx$. }
  \label{fig:spe}
\end{figure}
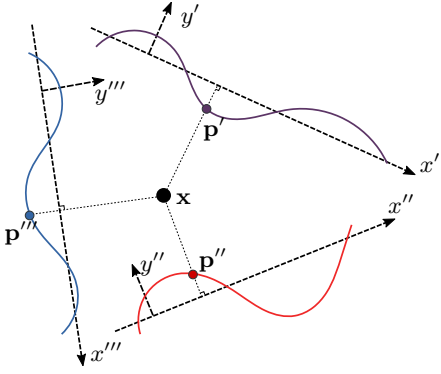

The key idea of our \textsc{Spe} is to use a set of parametric Spline functions as encoding functions. 
Different from \textsc{Fpe} which uses predefined sinusoidal functions, our \textsc{Spe} is trainable and optimized together with MLPs. 
In our implementation, we choose the B-Spline function due to its simplicity.

\paragraph{B-Spline function.}
We first briefly introduce the B-Spline basis and B-Spline functions. 
The B-Spline basis $B^i(x)$ is a locally supported function, where $B^i: \mathbb{R} \mapsto \mathbb{R}$ and $i$ is its polynomial degree. 
The $B^{0}(x)$ is defined as follows:
\begin{equation}
  B^0(x) =
    \begin{cases}
      1 & \text{if}\; |x| < 0.5; \\
      0 & \text{otherwise}.
    \end{cases}
\end{equation}
$B^i(x)$ is set to the $n^{th}$ convolution of $B^0(x)$. 
The linear B-Spline basis $B^1(x)$ is supported in $[-1, 1]$, and the quadratic B-Spline $B^2(x)$ is supported in $[-1.5, 1.5]$. 
And for simplicity, we use $B(x)$ and omit the superscript.

Given an input 1D domain, we first uniformly subdivide it to $K$ segments and get $(K+1)$ knot points $\{c_i\}_{i=0}^K$, and denote the interval between two knots as $\delta$. 
We scale and translate the B-Spline basis $B(x)$ to each knot point, and get $B_{\delta,c_i}(x) = B(\frac{x - c_i} {\delta})$. 
At the $i^{th}$ node point, we define an optimizable parameter $W_i$. 
The parametric Spline function is defined as
\begin{equation}
  \mathbf{\psi}(x) = \sum_{i=0}^K W_i B_{\delta,{c_i}}(x).
  \label{equ:spline}
\end{equation}
If we further define $W_i$ as a $C$-channel vector, we can obtain $C$ spline functions. 

\paragraph{Spline positional encoding.}
Without loss of generality, we assume the input domain for training MLP is $[-1, 1]^d$. 
We randomly select a set of unit directions $\mathbf{D}_1, \ldots, \mathbf{D}_m$, and these directions can determine a set of line segments with the same direction passing through the origin, whose two ends are on the unit sphere. 
On each line segment $L_k$, we can define a spline function $\mathbf{\psi}_{k}$ within the interval $[-1,1]$.
Given a point $\mx \in [-1,1]^d$, its spline positional encoding is defined as follows. 
We first compute the 1-D coordinate of $\mx$ with represent to each direction $\mathbf{D}_k$, denoted by $x_k$, by projecting it onto $L_k$:
\begin{equation}
  x_k := \langle \mx, \mathbf{D}_k \rangle.
\end{equation} 
The \textsc{Spe} of $\mx$ is defined as:
\begin{equation}
  \begin{aligned}
    \mathbf{\Phi}(\mx) = [\mathbf{\psi}_{1}(x_1), ... , \mathbf{\psi}_{M}(x_M)].
  \end{aligned}
  \label{equ:spe}
\end{equation}
To be able to differentiate different points in $\mathbb{R}^d$, the projection directions should be independent, and the projection direction number should be larger than $d$.

The above spline positional encoding lifts the point from a $d$-dimension vector up to a $C \times M$ tensor. 
In our experiments, we simply sum up the $M$ projections and get a $C$ dimension positional encoding.
The total number of parameters used by \textsc{Spe} is $C \times (K+1) \times M + (d-1) \times M $. 
Here $(d-1) \times M$ is for the projection directions.  
All the parameters are differentiable in \cref{equ:spe}, thus can be trained to find their optimal values.


\paragraph{Relationship with prior positional encodings.}
For an MLP that directly takes coordinates as input, we can define its positional encoding as $\phi(\mx)=\mx$. 
The Fourier positional encoding proposed by~\cite{Tancik2020,Mildenhall2020,Zhong2020} is composed of a set of sinusoidal functions with different frequencies, which can be defined as
\begin{align*}
  \Phi(\mx) = [ &\sin(2
  \pi w_1^T x), \cos(2 \pi w_1^T x), \cdots, \\ & \sin(2 \pi w_M^T x), \cos(2 \pi w_M^T
  x)].
\end{align*}
Since the spline function with sufficient knots can well approximate the Identity, Sine, and Cosine functions, our \textsc{Spe} can be regarded as a generalization of prior positional encodings. 
Actually, we can properly initialize  $W_i$ in \cref{equ:spline} according to \textsc{Fpe} and fix $W_i$ during the optimization process and achieve the same effect as $\textsc{Fpe}$.

\subsection{Training Scheme} \label{subsec:training}

\paragraph{Multi-scale optimization of \textsc{Spe}.} 
The B-Spline bases can be subdivided in a multi-scale manner, which is widely used in the multi-resolution optimization in Finite Element Analysis~\cite{Logan2017}. 
Suppose a Spline function is composed by $K$ \emph{linear} Spline bases, as defined in \cref{equ:spline}, we can refine it by subdividing the input domain to $2K$ segments and initialize the new weights $\hat{W}_i$ via the following formula:
\begin{equation}
  \hat{W}_j = \sum_{i=0}^K W_i B_{\delta,{c_i}}(\hat{c}_j)
  \label{equ:refine}
\end{equation}
where $\hat{c}_j$ represents the $j$-{th} refined knot. 
Other higher-order Spline bases can also be subdivided similarly, and we omit the detailed formulas for simplicity.
When training the network with the loss function, we first warm up the training process with a coarse resolution \textsc{Spe}. 
With a coarse \textsc{Spe}, the MLP quickly fits the low-frequency part of SDFs and provides a good initialization. 
Then we progressively refine \textsc{Spe} to increase the fitting ability of MLPs. 
In this way, our network can converge to better local minima: both the SDF away from the input points and the geometric details on the surface are better recovered. 

\begin{table*}[t]
  \tablestyle{5pt}{1.1}
  \scalebox{0.9}{
  \begin{tabular}{l|rr|rr|rr|rr|rr|rr|rr}
    \toprule
    Model          &  \multicolumn{2}{c|}{Armadillo} & \multicolumn{2}{c|}{Bimba} & \multicolumn{2}{c|}{Bunny} & \multicolumn{2}{c|}{Dragon} & \multicolumn{2}{c|}{Fandisk} & \multicolumn{2}{c|}{Gargogle} & \multicolumn{2}{c}{Dfaust} \\
                   &  \small Chamfer &  \small  MAE   &  \small Chamfer &  \small  MAE  &  \small Chamfer &  \small  MAE  &  \small Chamfer &  \small  MAE  &  \small Chamfer &  \small  MAE  &  \small Chamfer &  \small  MAE  &  \small Chamfer &  \small  MAE \\
    \midrule
    \textsc{Igr}   &  13.6 & \textbf{1.9}    &    5.1  &  1.1   &   2.5 &  \textbf{0.7}  &  62.1  &  1.8  &    2.3  &  1.0  &  17.2  &  6.1  &  17.6  &  \textbf{1.4} \\
    \textsc{Siren} &   2.2 & 22.1   &    5.6  & 18.9   &   1.5 & 16.3  &   \textbf{1.4}  &  2.3  &  243.3  & 20.3  &   2.8  & 17.0  &   9.3  & 32.9 \\
    \textsc{Fpe}   & 207.5 & 28.3   & 3867.2  & 27.1   & 263.7 & 25.4  & 528.2  & 30.3  & 6956.8  & 27.7  & 7342.5 & 24.9  & 116.7  & 38.5 \\
    \textsc{Spe}   &   \textbf{1.3} & 3.1    &    \textbf{1.6}  &  \textbf{0.6}   &   \textbf{1.5} &  0.8  &  1.8  &  \textbf{2.1}  &    \textbf{1.3}  &  \textbf{0.4}  &    \textbf{2.4} &  \textbf{1.3}  &   \textbf{9.1}  &  2.0 \\
    \bottomrule
  \end{tabular}
  } 
  \caption{Numerical results on SDF reconstruction from unorganized point
  clouds. The Chamfer distance and MAE are multiplied by $10000$ and $100$.
  Our \textsc{Spe} has much lower  Chamfer distance than \textsc{Fpe} and \textsc{Igr}, and better MAE than \textsc{Siren} and \textsc{Fpe}.}
  \label{tab:results}
\end{table*}

\paragraph{Network training.}
By default, we use an MLP with 4 fully-connected (FC) layers with the Softplus activation function, each of which contains 256 hidden unit, and choose linear B-Spline bases for \textsc{Spe}. 
In each iteration during the training stage, we randomly sample 10k to 20k points from the input point cloud and the same number of random points from the 3D bounding box containing the shape.
All input points are encoded via our \textsc{Spe}. 
We set the parameters of \textsc{Spe} to $K=256, C=64, M=3$, resulting a 64 dimension encoding for each point. 
As a reference, with \textsc{Fpe}, the dimension of per-point encoding is 256. The encoded point features are forwarded by the MLP. 
Then the loss in \cref{equ:loss} is calculated.
The parameters $\lambda$ and $\tau$ in \cref{equ:loss} are set to $0.1$ and $1$.
The MLP and \textsc{Spe} are optimized via the Adam~\cite{Kingma2014a} solver with a learning rate of $0.0001$,  without using the weight decay and normalization techniques.

For the multiscale optimization, we first initialize \textsc{Spe} with $K=2$, then progressively increase $K$ to 8, 32, 128, and 256, with the initialization method provided in \cref{equ:refine}.
When optimizing MLPs with $K=2$, we occasionally observe the extracted surface containing spurious patches away from the input point cloud. 
Inspired by the geometric initialization proposed by~\cite{Atzmon2020} which initializes the network to approximate a sphere, we train a randomly initialized MLP to fit the SDF of a sphere.
After training, the network weights are stored and used as the initialization of MLPs with $K=2$ in \textsc{Spe}.

For learning shape spaces, we train an Auto-Decoder proposed by~\cite{Park2019}.
Instead of relying on a global shape code to identity each shape~\cite{Park2019,Gropp2020}, our \textsc{Spe} itself is optimizable for each shape, which can be directly used to distinguish different shapes.
Therefore, we train a shared MLP and specific \textsc{Spe} for each shape in the training set.
The MLP is also composed of 4 FCs with 256 hidden units.
After training, the network weights are fixed, and only the \textsc{Spe} is optimized to fit new shapes in the testing set.
\section{Experiments and Evaluation}
\label{sec:result}
We have conducted the comparisons with several state-of-the-art methods to verify the effectiveness of our method.
Specifically, we regard the MLP that directly takes the coordinates as input as the baseline, \ie, \textsc{Igr}~\cite{Gropp2020}.
For the positional encoding, we compare our \textsc{Spe} with \textsc{Fpe} proposed by~\cite{Tancik2020}, which is an enhanced and improved version of the positional encoding in~\cite{Mildenhall2020,Zhong2020}, and \textsc{Siren}~\cite{Sitzmann2020}.
By default, these networks are all composed of 4 FC layers with 256 hidden units.

Our implementation is based on PyTorch, and all experiments were done with a desktop PC with an Intel Core i7 CPU (3.6 GHz) and GeForce 2080 Ti GPU (11 GB memory).
Our code and trained models are available at \url{https://wang-ps.github.io/spe}.

\begin{figure}[t]
  \centering
  \begin{overpic}[width=1\linewidth]{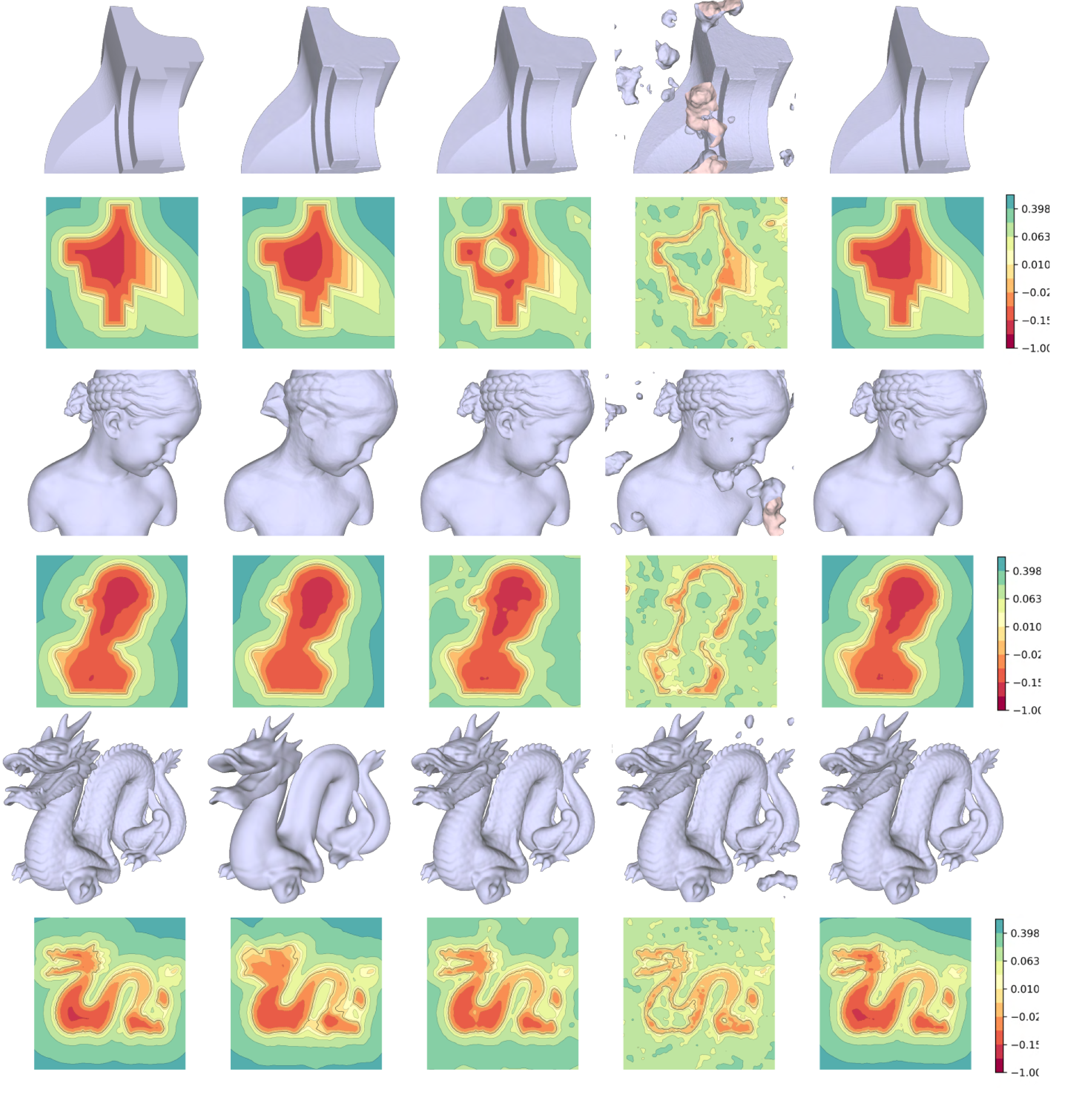}
    \put(7,  0){\small  \textsc{Gt}}
    \put(22, 0){\small (a) \textsc{Igr}}
    \put(38, 0){\small (b) \textsc{Siren}}
    \put(58, 0){\small (c) \textsc{Fpe}}
    \put(75, 0){\small (d) \textsc{Spe}}
  \end{overpic}
  \caption{Visual comparisons on SDF reconstruction from raw point clouds. 
  The reconstructed shapes and SDF slices are illustrated.}
  \label{fig:recons}
\end{figure}
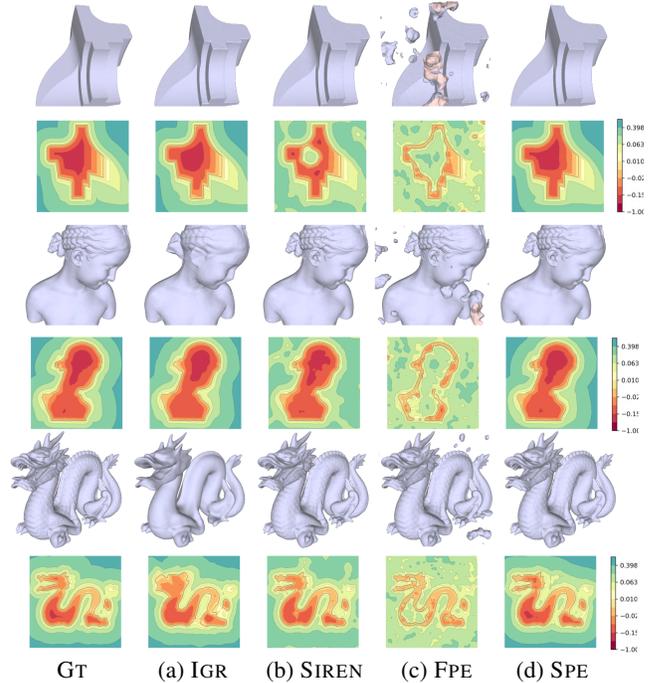

\subsection{Single Shape Learning} \label{sec:shape}
In this section, we test our method on the task of reconstructing SDFs from raw point clouds.
We evaluate both the quality of the reconstructed surface and SDFs.

\paragraph{Dataset.}
We collect 7 3D shapes as the benchmark, which include detailed geometric textures (Bunny, Armadillo, and Gargoyle), smooth surfaces (Bimba and Dragon), and sharp features (Fandisk).
The Dfaust point cloud is produced by a real scanner provided by~\cite{Bogo2017}.
For other models, we sample points with normals from the surface via uniformly placing virtual depth cameras around each shape.

\paragraph{Evaluation metric.}
We use the Chamfer distance to measure the quality of the extracted surface.
Specifically, we randomly sample a set of $N$ points $\mathcal{X} = \{x_i\}_{i=1}^N$ from the extracted surface and ground-truth surface $\hat{\mathcal{X}} = \{\hat{x}_i\}_{i=1}^N$, where $N=25k$.
And we use the following formula to calculate the Chamfer distance:
\begin{equation}
D(\mathcal{X},\hat{\mathcal{X}})
        = \frac{1}{N} \sum_i \min_j \|x_i - \hat{x}_j\| +
          \frac{1}{N} \sum_j \min_i \|\hat{x}_i - x_j\|
\label{equ:chamfer}
\end{equation}
We use the mean absolute error (MAE) between the predicted and ground-truth SDFs to measure the quality of the predicted SDFs.
To calculate the MAE, we uniformly draw $256^3$ samples on both the predicted and ground-truth SDFs.

\begin{figure}
  \centering
  \begin{overpic}[width=1\linewidth]{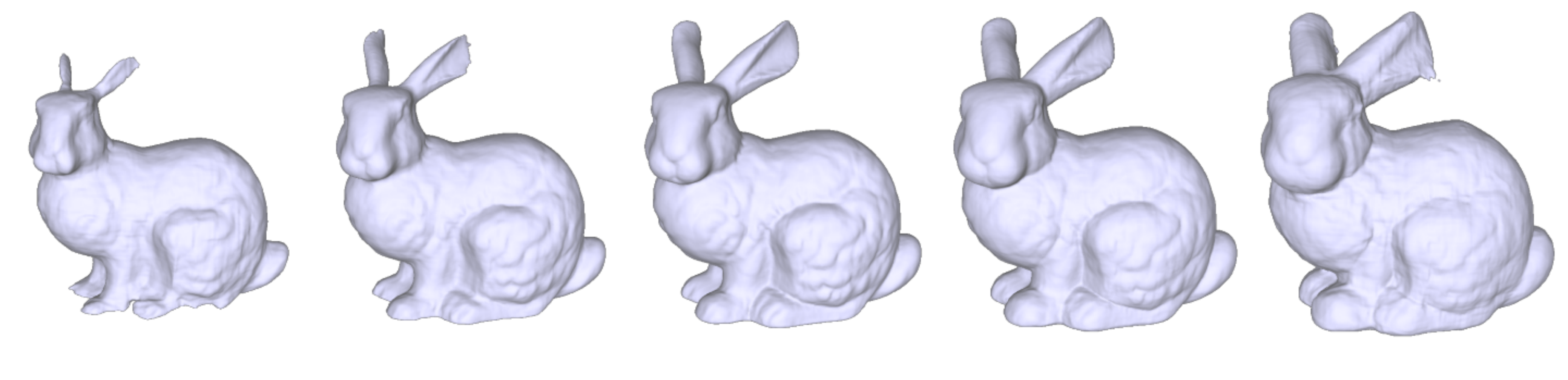}
    \put(7,  0){\small  -0.1}
    \put(28, 0){\small  -0.05}
    \put(50, 0){\small  0}
    \put(68, 0){\small  0.05}
    \put(90, 0){\small  0.1}
  \end{overpic}
  \caption{Different level sets of trained SDFs by \textsc{Spe}.
  }
  \label{fig:sdf}
\end{figure}
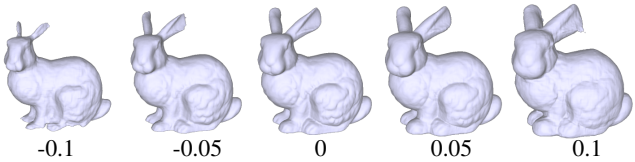

\paragraph{Results.}
The numerical results are summarized in \cref{tab:results}, and the visual results are shown in  \cref{fig:recons}.
As we can see, compared with \textsc{Igr}, the Chamfer distance of our method is greatly reduced, and the high-frequency geometric details are reconstructed much better, which verifies that with our \textsc{Spe} MLPs can easily reproduce the high-frequency details.
For \textsc{Siren} and \textsc{Fpe}, their extracted surfaces may contain spurious components, as shown in \cref{fig:recons}.
Note that it is non-trivial to fix this issue for their results: although the isolated small meshes can be easily removed, the incorrect components attached to the real surface are hard to remove and repair.
Moreover, the implicit fields of \textsc{Fpe} and \textsc{Siren} have large deviations from the ground-truth SDFs, as revealed by significantly larger MAE. 
In \cref{fig:sdf}, we show an application using SDFs trained with our \textsc{Spe} to extract the different level sets for shape shrinking and dilation.

\subsection{Ablation Study}

\paragraph{Expressiveness of \textsc{Spe}.}
We did an ablation study on how the choices of hyper-parameters (the segmentation number $K$, the channel of weights $C$, the projection number $M$, and the order of B-Spline basis), affect the performance of \textsc{Spe}. 
The experiments were done on the reconstruction of the Bimba model and we increased one hyper-parameter while keeping others unchanged.
The baseline is $K=128,C=64,M=3$ with linear B-Spline bases. 
The results summarized in \cref{tab:expressiveness} show that larger hyper-parameters can result in better apprimation quality,  while the segmentation number brings the most effective improvements.

\begin{table}[t]
  \tablestyle{2pt}{1.1}
  \scalebox{0.85}{
  \begin{tabular}{lccccc}
    \toprule
    (K, C, M)      &  (128, 64, 3) & (\textbf{256}, 64, 3) & (128, \textbf{128}, 3) & (128, 64, \textbf{6}) & (128, 64, 3)$^\star$\\
    \midrule
    Chamfer        &  1.72         & 1.63         &  1.67         &  1.69         & 1.71 \\
    \bottomrule
  \end{tabular}
  } 
  \caption{
  Chamfer distances on the Bimba model with different parameter settings.
  (128, 64, 3)$^\star$ uses quadratic B-Spline bases.
  }
  \label{tab:expressiveness}
\end{table}

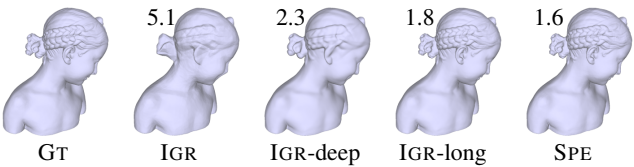
\begin{figure}[b]
  \centering
  \begin{overpic}[width=\linewidth]{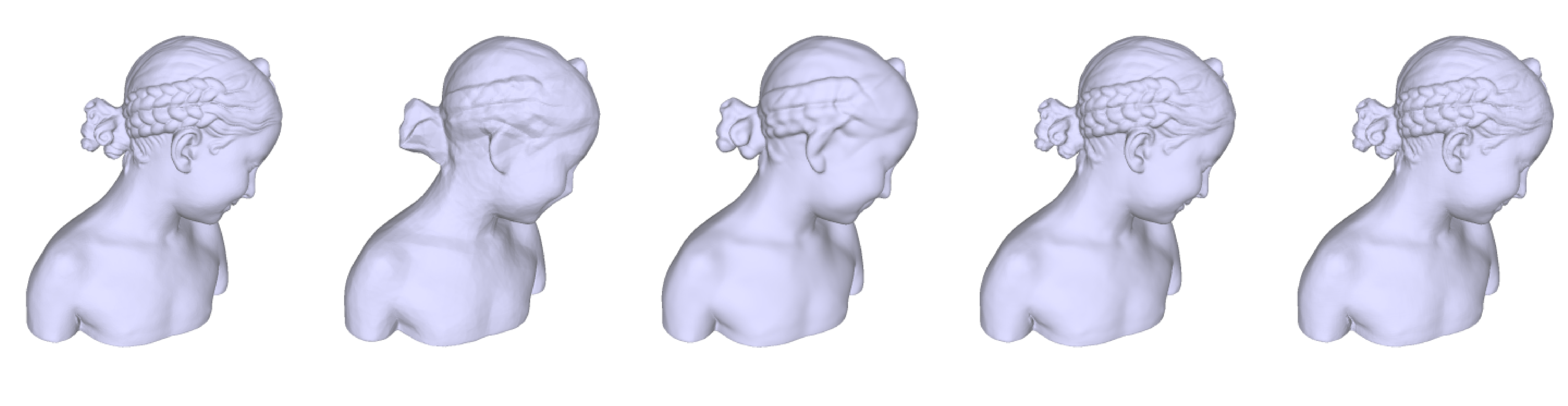}
    \put(7,  0){\small \textsc{Gt}}
    \put(26, 0){\small \textsc{Igr}}
    \put(24, 21) {\small 5.1}
    \put(43, 0){\small \textsc{Igr}-deep}
    \put(44, 21) {\small 2.3}
    \put(63, 0){\small \textsc{Igr}-long}
    \put(64, 21) {\small 1.8}
    \put(87, 0){\small  \textsc{Spe}}
    \put(84, 21) {\small 1.6}
  \end{overpic}
  \caption{
    Test on vanilla MLP with different settings. From left to right: the ground-truth, the result of \textsc{Igr} as the baseline,  \textsc{Igr}-deep: \textsc{Igr} with 4 times increased network depth, \textsc{Igr}-long: \textsc{Igr} with 10 times longer training time, and our \textsc{Spe} result.  The numbers in the figure are the Chamfer distance. }
  \label{fig:mlp-igr}
\end{figure}

\paragraph{Can a vanilla MLP compete with \textsc{Spe}?} 
To check whether a vanilla MLP can fit the high-frequency details, we train vanilla MLP -\textsc{Igr} by increasing its network depth and training time on the task of reconstructing the Bimba model.
The results are shown in \cref{fig:mlp-igr}. As we can see, even by increasing the network depth by 4 times or increasing the training time by 10 times, the results of vanilla MLPs are still worse than \textsc{Spe}. Without any kind of positional encoding, the vanilla MLP converges too slow to recover the fine details.

\paragraph{Specialization of our \textsc{Spe}.}
With proper initialization, our \textsc{Spe} can reproduce the MLP with \textsc{Fpe}, since B-Spline functions can  fit sinusoidal functions well with sufficiently small $\delta$.
In practice, we find that as long as the $\delta$ of B-Spline functions is similar to the period of a sinusoidal function, we can achieve similar effects.

\begin{table}[t]
  \centering
  \scalebox{0.9}{
    \begin{tabular}{c|ccccc}
      \toprule
      Method & \textsc{Igr} & \textsc{Siren} & \textsc{Fpe} & \textsc{Spe} \\
      \midrule
      Chamfer  &14.1  & 15.3  & 18.1 & \textbf{11.5}   \\
      MAE      &3.6   & 34.7  & 30.4 & \textbf{3.4} \\
      \bottomrule
    \end{tabular}
  } 
  \caption{Comparisons of shape space learning on the D-Faust dataset.}
  \label{tab:space}
\end{table}

\begin{figure}[t]
  \centering
  \begin{overpic}[width=1\linewidth]{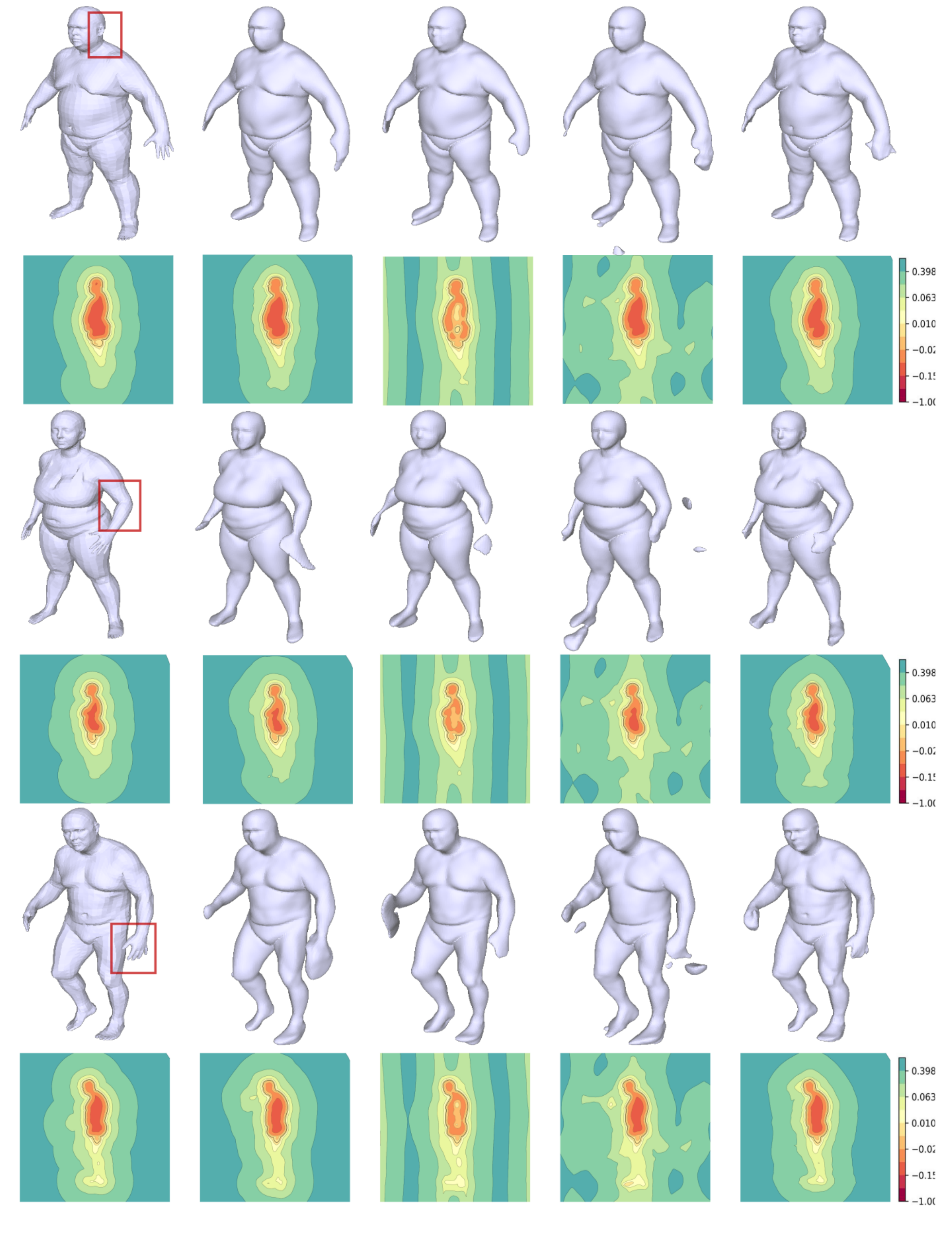}
    \put(6,  0){\small  \textsc{Gt}}
    \put(18, 0){\small (a) \textsc{Igr}}
    \put(31, 0){\small (b) \textsc{Siren}}
    \put(48, 0){\small (c) \textsc{Fpe}}
    \put(63, 0){\small (d) \textsc{Spe}}
  \end{overpic}
  \caption{Visual comparisons on shape space learning. 
  3 human body shapes from the testing set reconstructed by different methods and the corresponding SDF slices are shown.}
  \label{fig:space}
\end{figure}
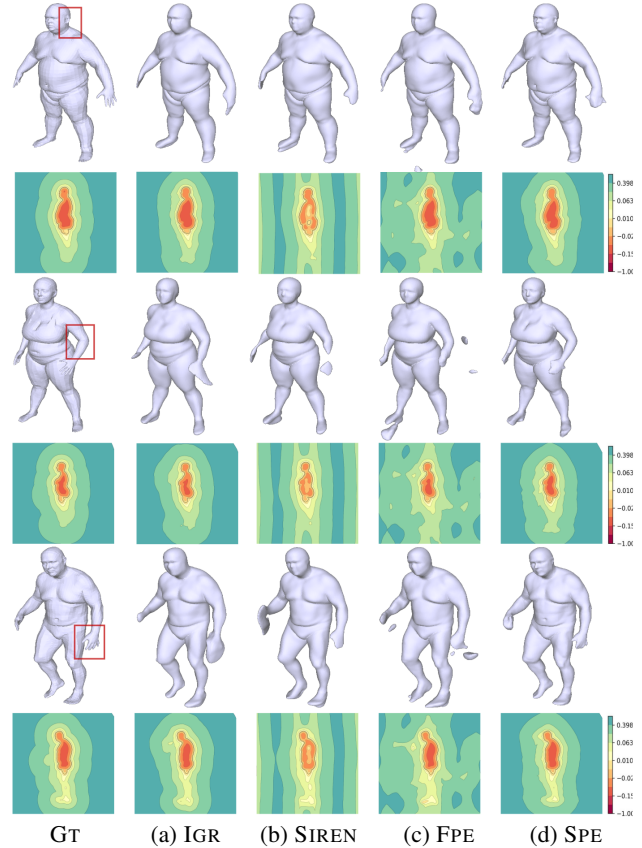

\subsection{Shape Space Learning}
\textsc{Spe} is also suitable to learn shape spaces from raw scans using auto-decoders~\cite{Park2019}. We compared our method with \textsc{Igr}, \textsc{Siren}, and \textsc{Fpe}, for which each shape has a specific shape code with dimension 256 and all the shapes share the same decoder.  Our \textsc{Spe} optimizes the specific spline positional encoding for each shape and does not use a shape code. The parameters are set to $K=64, C=32, M=3$.
We use the Chamfer distance and MAE defined in \cref{subsec:learnsdf} as the evaluation metrics to compare the quality of the reconstructed surface and SDFs on the unseen testing shapes.

\paragraph{Dataset.}
We conducted the experiment on the D-Faust dataset~\cite{Bogo2017} by following the setup of \cite{Gropp2020}.
D-Faust contains real scans of human bodies in different poses, in which 6258 shapes are used for training and 181 shapes are used for testing.
For each raw scans, we sample 250k points with normals as the input.

\paragraph{Results.}
After training, all the methods are tested on the test shapes by fixing the network weights and optimizing the shape code or our spline weights.
\cref{tab:space} lists the Chamfer distance and the MAE of SDFs of the results generated by four methods.
It is clear that our method outperforms the other three methods, and the second-best method is \textsc{Igr}.
Although \textsc{Siren} and \textsc{Fpe} have a better fitting ability in single shape reconstruction, they are even worse than \textsc{Igr} in the shape space learning within the auto-decoder framework. 
The visual comparisons in \cref{fig:space} further confirm the better performance of our method.

\subsection{Generalizability of \textsc{Spe}}
In this section, we show the generalizability of our method via the tasks of
fitting images and SDFs.

\begin{figure}[t]
  \centering
  \begin{overpic}[width=1\linewidth]{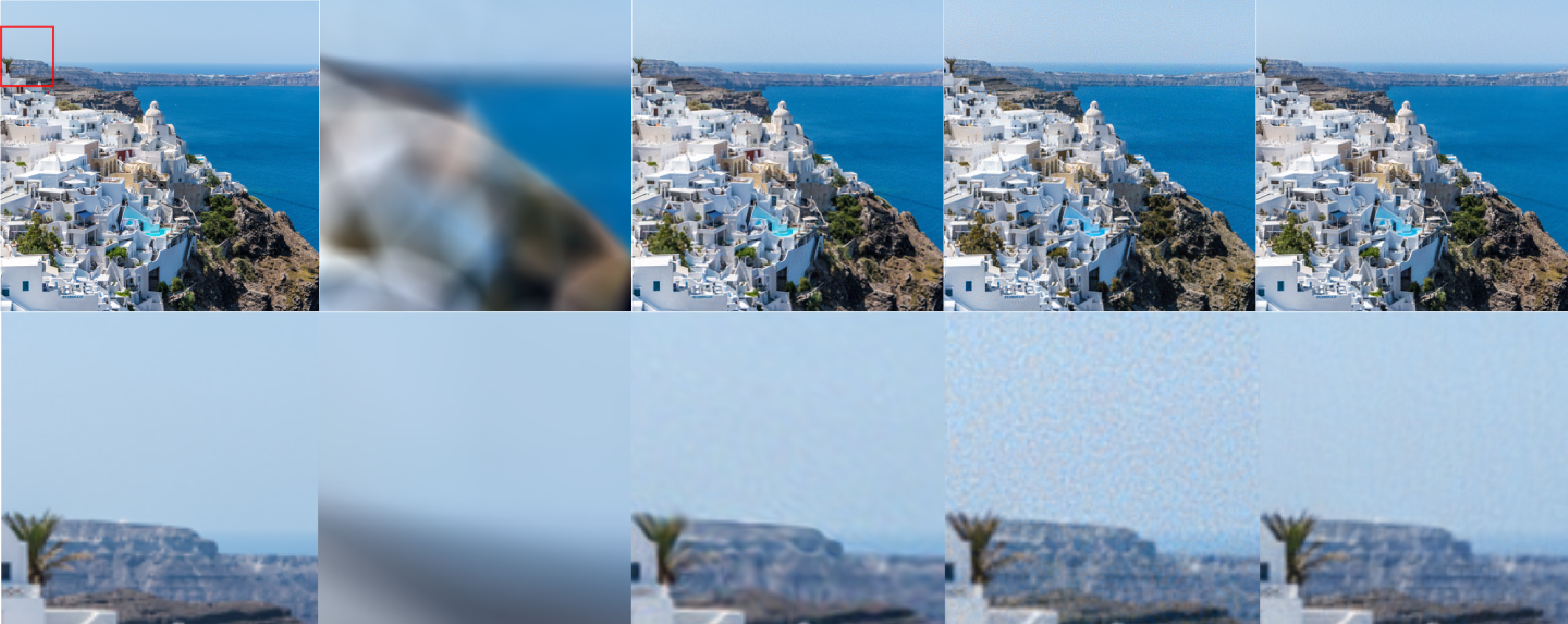}
    \put(7, -4){\small  \textsc{Gt}}
    \put(23, -4){\small (a) \textsc{Mlp}}
    \put(42, -4){\small (b) \textsc{Siren}}
    \put(64, -4){\small (c) \textsc{Fpe}}
    \put(84, -4){\small (d) \textsc{Spe}$^\star$}
  \end{overpic}
  \vspace{0.01mm}
  \caption{Results of image reconstruction. The bottom row shows zoom-in views.}
  \label{fig:image}
\end{figure}
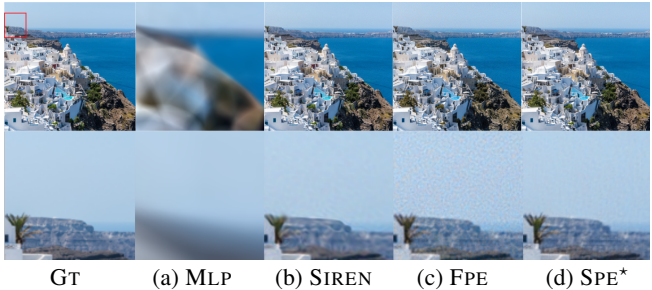

\paragraph{Image fitting.}
In this experiment, we trained an MLP to map 2D coordinates to the corresponding pixel values. 
We use the image dataset provided by~\cite{Tancik2020}, which contains 16 natural images and 16 text images. 
The MLP is trained with an $L_2$ loss. We use PSNR as the evaluation metric.
The results are summarized in \cref{tab:img}.
With similar size of parameters, our \textsc{Spe} achieves comparable results to \textsc{Siren} and \textsc{Fpe}, and is much better than a vanilla \textsc{Mlp}.
With more projection directions ($M=32$), the performance of our \textsc{Spe} can be significantly improved. \cref{fig:image} illustrates the reconstruction results and zoom-in views by different methods. We notice that  the resulting images of \textsc{Fpe} have visible noise patterns, as shown by the zoom-in figure, and the result of Siren is more blur than ours.

\begin{table}[t]
  \centering
  \scalebox{0.9}{
  \begin{tabular}{c|ccccc}
    \toprule
    Method & \textsc{Mlp} & \textsc{Siren} & \textsc{Fpe} & \textsc{Spe} & \textsc{Spe}$^\star$  \\
    \midrule
    Natural Images  & 18.3  & 31.1  & 30.8 & 30.1 & \textbf{33.6} \\
    Text    Images  & 18.4  & 35.6  & 33.7 & 37.4 & \textbf{40.4} \\
    \bottomrule
  \end{tabular}
  } 
  \caption{PSNRs of the image reconstruction task. \textsc{Spe}$^\star$ is \textsc{Spe} with $M=32$. }
  \label{tab:img}
\end{table}

\begin{table}[t]
  \centering
  \scalebox{0.9}{
  \begin{tabular}{ccccccc}
    \toprule
    Model          &  Arm. & Bimba & Bunny & Dragon & Fandisk & Garg. \\
    \midrule
    \textsc{Mlp}   &  9.69 & 4.69   & 3.76  & 11.59  &   2.06  &  9.94  \\
    \textsc{Siren} &  1.19 & 1.41   & 1.48  &  1.23  &   1.44  &  \textbf{1.66}  \\
    \textsc{Fpe}   &  1.24 & 1.44   & 1.49  &  1.29  &   1.47  &  1.71  \\
    \textsc{Spe}   &  \textbf{1.18} & \textbf{1.39}   & \textbf{1.46}  &  \textbf{1.19}  &   \textbf{1.36}  &  1.69   \\
    \bottomrule
  \end{tabular}
  } 
  \caption{Comparisons on the SDF regression task. The numbers are Chamfer distances.}
  \label{tab:regression}
\end{table}

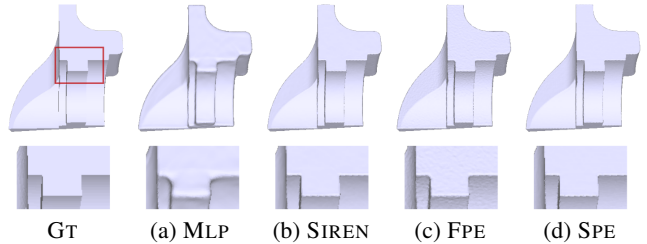
\begin{figure}[t]
  \centering
  \begin{overpic}[width=1\linewidth]{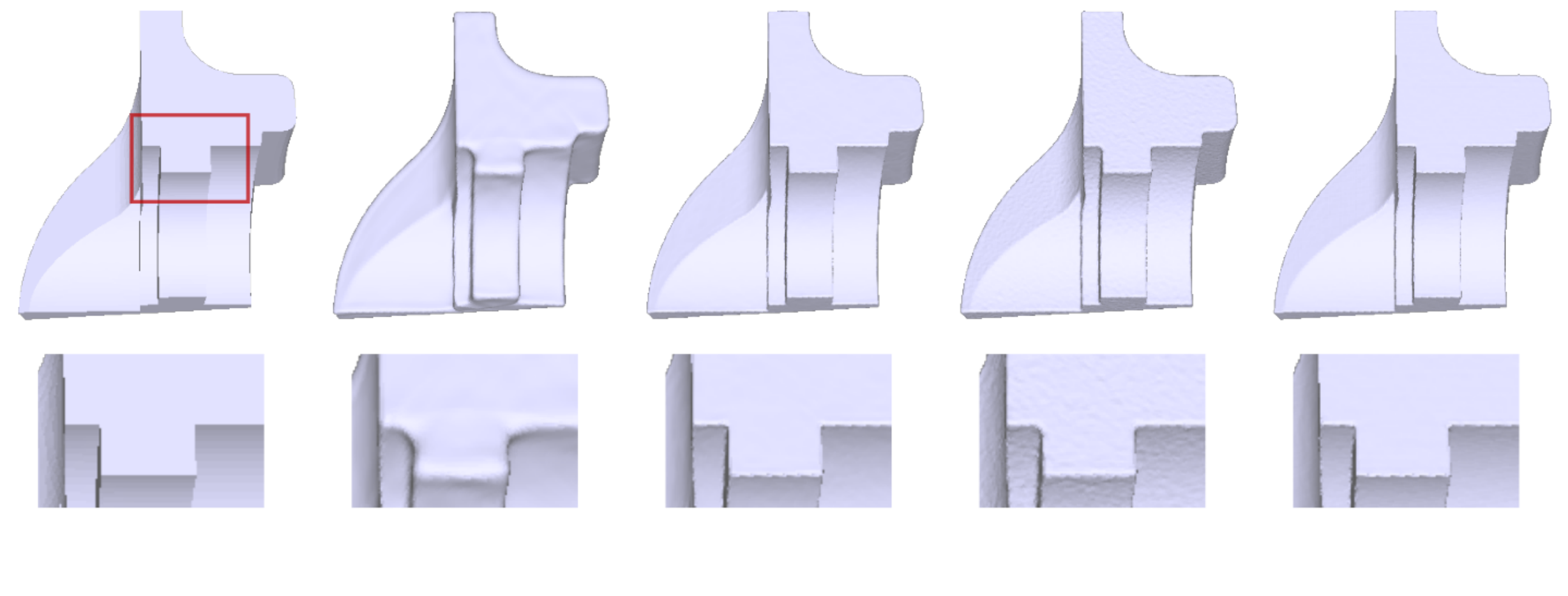}
    \put(7,  1){\small  \textsc{Gt}}
    \put(23, 1){\small (a) \textsc{Mlp}}
    \put(42, 1){\small (b) \textsc{Siren}}
    \put(64, 1){\small (c) \textsc{Fpe}}
    \put(84, 1){\small (d) \textsc{Spe}}
  \end{overpic}
  \caption{Comparisons on shape regression. The results are shown in the first row, and the zoom-in figures are shown in the second row.}
  \label{fig:sdfregress}
\end{figure}

\paragraph{SDF fitting.}
Instead of learning SDFs from raw points, we sample the ground-truth SDFs in the resolution of $256^3$. 
The MLP takes 3D coordinates as input and output 1D distance values and directly fits the ground-truth SDFs with an $L_1$ loss~\cite{Park2019}. 
We set the projection direction of our \textsc{Spe} as $16$ to get smoother results.
We use the same dataset in \cref{sec:shape} and summarize the fitting results in \cref{tab:regression}. The zero isosurface results of the Fandisk example by different methods are illustrated in \cref{fig:sdfregress}.  It can be seen that the results of \textsc{Spe} and \textsc{Siren} are more faithful to the ground truth than other methods, while the result of \textsc{Igr} is over-smoothed and  the result of \textsc{Fpe} has visible noise patterns.



\section{Conclusion} \label{sec:conclusion}
We present a novel and effective Spline positional encoding scheme for learning 3D implicit signed distance fields from raw point clouds.
The spline positional encoding enhances the representation power of MLPs and outperforms the existing positional encoding schemes like Fourier positional encoding and \textsc{Siren} in recovering SDFs.

In the future, we would like to explore \textsc{Spe} in the following directions.

\paragraph{Non-uniform spline knots.}
Compared with uniform knots we used in spline functions, non-uniform knots provide more freedom to model complex and non-smooth spline functions and would also help reduce the parameter sizes of \textsc{Spe} while keeping the same approximation power.

\paragraph{Composition of positional encoding.}
As positional encoding has proved to be an effective way to distinguish nearby points in a higher dimension space, it would be interesting to composite multiple scale \textsc{Spe}s to strengthen the capability of \textsc{Spe} while using fewer parameters for each \textsc{Spe}.

\section*{Acknowledgements}
We wish to thank the anonymous reviewers for their valuable feedback.

\clearpage
\bibliographystyle{named}
\bibliography{src/ref/reference}

\begin{thebibliography}{}

\bibitem[\protect\citeauthoryear{Atzmon and Lipman}{2020a}]{Atzmon2020}
Matan Atzmon and Yaron Lipman.
\newblock {SAL}: Sign agnostic learning of shapes from raw data.
\newblock In {\em CVPR}, 2020.

\bibitem[\protect\citeauthoryear{Atzmon and Lipman}{2020b}]{Atzmon2020a}
Matan Atzmon and Yaron Lipman.
\newblock {SAL++}: Sign agnostic learning with derivatives.
\newblock {\em arXiv preprint arXiv:2006.05400}, 2020.

\bibitem[\protect\citeauthoryear{Bogo \bgroup \em et al.\egroup
  }{2017}]{Bogo2017}
Federica Bogo, Javier Romero, Gerard Pons-Moll, and Michael~J Black.
\newblock Dynamic {FAUST}: Registering human bodies in motion.
\newblock In {\em CVPR}, 2017.

\bibitem[\protect\citeauthoryear{Bridson}{2015}]{Bridson2015}
Robert Bridson.
\newblock {\em Fluid simulation for computer graphics}.
\newblock CRC press, 2015.

\bibitem[\protect\citeauthoryear{Chabra \bgroup \em et al.\egroup
  }{2020}]{Chabra2020}
Rohan Chabra, Jan~Eric Lenssen, Eddy Ilg, Tanner Schmidt, Julian Straub, Steven
  Lovegrove, and Richard Newcombe.
\newblock Deep local shapes: Learning local {SDF} priors for detailed {3D}
  reconstruction.
\newblock In {\em ECCV}, 2020.

\bibitem[\protect\citeauthoryear{Chen and Zhang}{2019}]{Chen2019}
Zhiqin Chen and Hao Zhang.
\newblock Learning implicit fields for generative shape modeling.
\newblock In {\em CVPR}, 2019.

\bibitem[\protect\citeauthoryear{Curless and Levoy}{1996}]{Curless1996}
Brian Curless and Marc Levoy.
\newblock A volumetric method for building complex models from range images.
\newblock In {\em SIGGRAPH}, 1996.

\bibitem[\protect\citeauthoryear{Genova \bgroup \em et al.\egroup
  }{2020}]{Genova2020}
Kyle Genova, Forrester Cole, Avneesh Sud, Aaron Sarna, and Thomas Funkhouser.
\newblock Local deep implicit functions for {3D} shape.
\newblock In {\em CVPR}, 2020.

\bibitem[\protect\citeauthoryear{Gropp \bgroup \em et al.\egroup
  }{2020}]{Gropp2020}
Amos Gropp, Lior Yariv, Niv Haim, Matan Atzmon, and Yaron Lipman.
\newblock Implicit geometric regularization for learning shapes.
\newblock In {\em ICML}, 2020.

\bibitem[\protect\citeauthoryear{Hart}{1996}]{Hart1996}
John~C Hart.
\newblock Sphere tracing: A geometric method for the antialiased ray tracing of
  implicit surfaces.
\newblock {\em The Visual Computer}, 12(10), 1996.

\bibitem[\protect\citeauthoryear{Jacot \bgroup \em et al.\egroup
  }{2018}]{Jacot2018}
Arthur Jacot, Franck Gabriel, and Cl{\'e}ment Hongler.
\newblock Neural tangent kernel: Convergence and generalization in neural
  networks.
\newblock In {\em NeurIPS}, 2018.

\bibitem[\protect\citeauthoryear{Jiang \bgroup \em et al.\egroup
  }{2020}]{Jiang2020}
Chiyu Jiang, Avneesh Sud, Ameesh Makadia, Jingwei Huang, Matthias Nie{\ss}ner,
  and Thomas Funkhouser.
\newblock Local implicit grid representations for {3D} scenes.
\newblock In {\em CVPR}, 2020.

\bibitem[\protect\citeauthoryear{Jones \bgroup \em et al.\egroup
  }{2006}]{Jones2006}
Mark~W Jones, J~Andreas Baerentzen, and Milos Sramek.
\newblock {3D} distance fields: A survey of techniques and applications.
\newblock {\em IEEE. T. Vis. Comput. Gr.}, 12(4), 2006.

\bibitem[\protect\citeauthoryear{Kingma and Ba}{2014}]{Kingma2014a}
Diederik~P. Kingma and Jimmy Ba.
\newblock Adam: A method for stochastic optimization.
\newblock {\em ICLR}, 2014.

\bibitem[\protect\citeauthoryear{Logan}{2017}]{Logan2017}
Daryl~L Logan.
\newblock {\em A first course in the finite element method}.
\newblock Nelson Education, 2017.

\bibitem[\protect\citeauthoryear{Lorensen and Cline}{1987}]{Lorensen1987}
William~E. Lorensen and Harvey~E. Cline.
\newblock {Marching Cubes}: A high resolution {3D} surface construction
  algorithm.
\newblock In {\em SIGGRAPH}, 1987.

\bibitem[\protect\citeauthoryear{Mescheder \bgroup \em et al.\egroup
  }{2019}]{Mescheder2019}
Lars Mescheder, Michael Oechsle, Michael Niemeyer, Sebastian Nowozin, and
  Andreas Geiger.
\newblock Occupancy networks: Learning {3D} reconstruction in function space.
\newblock In {\em CVPR}, 2019.

\bibitem[\protect\citeauthoryear{Mildenhall \bgroup \em et al.\egroup
  }{2020}]{Mildenhall2020}
Ben Mildenhall, Pratul~P Srinivasan, Matthew Tancik, Jonathan~T Barron, Ravi
  Ramamoorthi, and Ren Ng.
\newblock {NeRF}: Representing scenes as neural radiance fields for view
  synthesis.
\newblock In {\em ECCV}, 2020.

\bibitem[\protect\citeauthoryear{Oechsle \bgroup \em et al.\egroup
  }{2019}]{Oechsle2019}
Michael Oechsle, Lars Mescheder, Michael Niemeyer, Thilo Strauss, and Andreas
  Geiger.
\newblock Texture fields: Learning texture representations in function space.
\newblock In {\em ICCV}, 2019.

\bibitem[\protect\citeauthoryear{Park \bgroup \em et al.\egroup
  }{2019}]{Park2019}
Jeong~Joon Park, Peter Florence, Julian Straub, Richard Newcombe, and Steven
  Lovegrove.
\newblock {DeepSDF}: Learning continuous signed distance functions for shape
  representation.
\newblock In {\em CVPR}, 2019.

\bibitem[\protect\citeauthoryear{Peng \bgroup \em et al.\egroup
  }{2020}]{Peng2020}
Songyou Peng, Michael Niemeyer, Lars Mescheder, Marc Pollefeys, and Andreas
  Geiger.
\newblock Convolutional occupancy networks.
\newblock In {\em ECCV}, 2020.

\bibitem[\protect\citeauthoryear{Rahaman \bgroup \em et al.\egroup
  }{2019}]{Rahaman2019}
Nasim Rahaman, Aristide Baratin, Devansh Arpit, Felix Draxler, Min Lin, Fred
  Hamprecht, Yoshua Bengio, and Aaron Courville.
\newblock On the spectral bias of neural networks.
\newblock In {\em ICML}, 2019.

\bibitem[\protect\citeauthoryear{Rahimi and Recht}{2008}]{Rahimi2018}
Ali Rahimi and Benjamin Recht.
\newblock Random features for large-scale kernel machines.
\newblock In {\em NeurIPS}, 2008.

\bibitem[\protect\citeauthoryear{Sitzmann \bgroup \em et al.\egroup
  }{2019}]{Sitzmann2019}
Vincent Sitzmann, Michael Zollh{\"o}fer, and Gordon Wetzstein.
\newblock Scene representation networks: Continuous {3D}-structure-aware neural
  scene representations.
\newblock In {\em NeurIPS}, 2019.

\bibitem[\protect\citeauthoryear{Sitzmann \bgroup \em et al.\egroup
  }{2020}]{Sitzmann2020}
Vincent Sitzmann, Julien~NP Martel, Alexander~W Bergman, David~B Lindell, and
  Gordon Wetzstein.
\newblock Implicit neural representations with periodic activation functions.
\newblock In {\em NeurIPS}, 2020.

\bibitem[\protect\citeauthoryear{Tancik \bgroup \em et al.\egroup
  }{2020}]{Tancik2020}
Matthew Tancik, Pratul~P. Srinivasan, Ben Mildenhall, Sara Fridovich-Keil,
  Nithin Raghavan, Utkarsh Singhal, Ravi Ramamoorthi, Jonathan~T. Barron, and
  Ren Ng.
\newblock Fourier features let networks learn high frequency functions in low
  dimensional domains.
\newblock In {\em NeurIPS}, 2020.

\bibitem[\protect\citeauthoryear{Vaswani \bgroup \em et al.\egroup
  }{2017}]{Vaswani2017}
Ashish Vaswani, Noam Shazeer, Niki Parmar, Jakob Uszkoreit, Llion Jones,
  Aidan~N Gomez, {\L}ukasz Kaiser, and Illia Polosukhin.
\newblock Attention is all you need.
\newblock In {\em NeurIPS}, 2017.

\bibitem[\protect\citeauthoryear{Xu \bgroup \em et al.\egroup
  }{2019}]{xu2019training}
Zhi-Qin~John Xu, Yaoyu Zhang, and Yanyang Xiao.
\newblock Training behavior of deep neural network in frequency domain.
\newblock In {\em International Conference on Neural Information Processing},
  2019.

\bibitem[\protect\citeauthoryear{Zhong \bgroup \em et al.\egroup
  }{2020}]{Zhong2020}
Ellen~D Zhong, Tristan Bepler, Joseph~H Davis, and Bonnie Berger.
\newblock Reconstructing continuous distributions of {3D} protein structure
  from cryo-{EM} images.
\newblock In {\em ICLR}, 2020.

\end{thebibliography}

\end{document}